\documentclass{article}

\usepackage{microtype}
\usepackage{graphicx}
\usepackage{subcaption}
\usepackage{booktabs} % for professional tables

\usepackage{hyperref}

\usepackage[preprint]{icml2026}

\usepackage{amsmath}
\usepackage{amssymb}
\usepackage{mathtools}
\usepackage{amsthm}

\usepackage{multirow}

\usepackage[capitalize,noabbrev]{cleveref}

\theoremstyle{plain}

\theoremstyle{definition}

\theoremstyle{remark}

\usepackage[disable,textsize=tiny]{todonotes}

\icmltitlerunning{Evaluating Nerve Segmentation in Brachial Plexus Ultrasound}

\begin{document}

\twocolumn[
	\icmltitle{Evaluating Deep Learning-Based Nerve Segmentation in Brachial Plexus Ultrasound Under Realistic Data Constraints}

	% It is OKAY to include author information, even for blind submissions: the
	% style file will automatically remove it for you unless you've provided
	% the [accepted] option to the icml2026 package.

	% List of affiliations: The first argument should be a (short) identifier you
	% will use later to specify author affiliations Academic affiliations
	% should list Department, University, City, Region, Country Industry
	% affiliations should list Company, City, Region, Country

	% You can specify symbols, otherwise they are numbered in order. Ideally, you
	% should not use this facility. Affiliations will be numbered in order of
	% appearance and this is the preferred way.
	\icmlsetsymbol{equal}{*}
	\begin{icmlauthorlist}
		\icmlauthor{Dylan Yves}{kmutt}
		\icmlauthor{Khush Agarwal}{kmutt}
		\icmlauthor{Jonathan Hoyin Chan}{kmutt}
        \icmlauthor{Patcharapit Promoppatum}{mchkmutt}
        \icmlauthor{Aroonkamon Pattanasiricharoen}{pracharak}

	\end{icmlauthorlist}

	\icmlaffiliation{kmutt}{School of Information and Technology, King Mongkut's University of Technology Thonburi, Bangkok, Thailand}
	\icmlaffiliation{mchkmutt}{Department of Mechanical Engineering, King Mongkut's University of Technology Thonburi, Bangkok, Thailand}
    \icmlaffiliation{pracharak}{Department of Anesthesiology, Charoenkrung Pracharak Hospital, Bangkok, Thailand}

	\icmlcorrespondingauthor{Dylan Yves}{dylan.yves@mail.kmutt.ac.th}
	\icmlcorrespondingauthor{Khush Agarwal}{khush.agar@kmutt.ac.th}
	\icmlcorrespondingauthor{Jonathan Hoyin Chan}{jonathan@sit.kmutt.ac.th}

	\icmlkeywords{Deep Learning, Ultrasound, Nerve Segmentation, Brachial Plexus}

	\vskip 0.3in
]

% this must go after the closing bracket ] following \twocolumn[ ...

% This command actually creates the footnote in the first column listing the
% affiliations and the copyright notice. The command takes one argument, which
% is text to display at the start of the footnote. The \icmlEqualContribution
% command is standard text for equal contribution. Remove it (just {}) if you
% do not need this facility.

% Use ONE of the following lines. DO NOT remove the command.
% If you have no special notice, KEEP empty braces:
\printAffiliationsAndNotice{}
% Or, if applicable, use the standard equal contribution text:
% \printAffiliationsAndNotice{\icmlEqualContribution}

\begin{abstract}
	Accurate nerve localization is critical for the success of ultrasound-guided regional anesthesia, yet manual identification remains challenging due to low image contrast, speckle noise, and inter-patient anatomical variability. This study evaluates deep learning-based nerve segmentation in ultrasound images of the brachial plexus using a U-Net architecture, with a focus on how dataset composition and annotation strategy influence segmentation performance. We find that training on combined data from multiple ultrasound machines (SIEMENS ACUSON NX3 Elite and Philips EPIQ5) provides regularization benefits for lower-performing acquisition sources, though it does not surpass single-source training when matched to the target domain. Extending the task from binary nerve segmentation to multi-class supervision (artery, vein, nerve, muscle) results in decreased nerve-specific Dice scores, with performance drops ranging from 9\% to 61\% depending on dataset, likely due to class imbalance and boundary ambiguity. Additionally, we observe a moderate positive correlation between nerve size and segmentation accuracy (Pearson $r=0.587$, $p<0.001$), indicating that smaller nerves remain a primary challenge. These findings provide methodological guidance for developing robust ultrasound nerve segmentation systems under realistic clinical data constraints.
\end{abstract}

\section{Introduction}

Ultrasound-guided regional anesthesia has become the standard of care for upper limb surgeries, with the brachial plexus block being among the most commonly performed procedures~\cite{ref12, ref13}. The brachial plexus, a network of nerves originating from cervical roots C5--C8 and T1, provides motor and sensory innervation to the shoulder, arm, forearm, and hand~\cite{gray2015anatomy}. Successful blockade of this nerve bundle can provide effective anesthesia while avoiding the systemic effects of general anesthesia, resulting in reduced healthcare costs and improved perioperative outcomes~\cite{ref14}. Misidentifying the nerve location can lead to incomplete blocks or serious complications including nerve injury, vascular puncture, and pneumothorax~\cite{ref17}.

Ultrasound imaging has revolutionized regional anesthesia by enabling real-time visualization of nerves, blood vessels, and surrounding tissues without ionizing radiation~\cite{ultrasoundguided2010}. Nevertheless, ultrasound-based nerve identification presents substantial interpretation challenges. Images are affected by speckle noise from acoustic wave interference, acoustic shadowing from overlying structures, and low contrast between adjacent soft tissues~\cite{ref16}. The brachial plexus appears as a relatively small structure, often a bundle of hypoechoic circles with a surrounding hyperechoic sheath, occupying only a tiny fraction of the image, which leads to extreme class imbalance between nerve and background pixels~\cite{ding2022mallesnet}. Inter-patient anatomical variability further complicates identification, with patient characteristics such as obesity reducing ultrasound clarity~\cite{ref18}. These factors collectively explain why procedural success remains strongly dependent on operator experience~\cite{ref19}.

Recent advances in deep learning have demonstrated strong potential for automated medical image analysis~\cite{ref20}, with convolutional neural network (CNN) models like U-Net approaching expert-level performance in anatomical structure segmentation~\cite{ref3}. For brachial plexus segmentation specifically, recent clinical trials have demonstrated that CNN-assisted identification can match experienced anesthesiologists~\cite{xi2024cnn, wang2024bpsegsys}. However, significant challenges remain. Most existing approaches rely on large, densely annotated datasets that are difficult to obtain in clinical practice, where expert annotation time is limited. Furthermore, models trained on data from one ultrasound machine often generalize poorly to images from different devices, limiting practical deployment~\cite{wang2024bpsegsys}. The effect of dataset composition, particularly combining data from heterogeneous sources, and annotation strategy on segmentation performance remains insufficiently explored.

This study addresses these gaps by evaluating deep learning-based nerve segmentation under realistic clinical data constraints. Using a public dataset of brachial plexus ultrasound images from two different machines, we investigate: (1) whether combining data from multiple ultrasound devices improves model generalization; (2) how multi-class annotation (nerve, artery, vein, muscle) compares to binary nerve-only supervision; and (3) the relationship between anatomical characteristics (nerve size) and segmentation accuracy. Our findings provide methodological guidance for developing robust ultrasound nerve segmentation systems when dataset diversity is limited and annotation resources are constrained.

\section{Literature Review}

\subsection{Deep Learning for Medical Image Segmentation}

Deep learning methods, particularly convolutional neural networks (CNNs), have transformed medical image segmentation, enabling automated delineation of anatomical structures including tumors, organs, and blood vessels~\cite{ref1}. Medical image segmentation techniques fall into two principal categories: semantic segmentation, which assigns each pixel to predefined classes, and instance segmentation (e.g., Mask R-CNN~\cite{he2017maskrcnn}, PANet~\cite{ref8}), which additionally distinguishes individual instances within the same class~\cite{ref2}. For ultrasound nerve segmentation, where the objective is to identify the nerve region rather than separate multiple nerve instances, semantic segmentation is the dominant approach.

The U-Net architecture introduced by Ronneberger et al.~\cite{ref3} has become the de facto standard for biomedical image segmentation due to its elegant encoder-decoder design with skip connections. The contracting encoder path captures multi-scale contextual features through successive convolution and pooling operations, while the expanding decoder path recovers spatial detail through upsampling and concatenation with corresponding encoder features. This design is particularly well-suited for segmenting small anatomical structures because it effectively combines global context with local boundary information. U-Net's ability to achieve strong performance with relatively limited training data, a crucial advantage given the scarcity of medical imaging datasets, has led to its widespread adoption. Other notable semantic segmentation architectures include FCN~\cite{ref4}, PSPNet~\cite{ref5}, and DeepLab~\cite{ref6}, though U-Net and its variants remain the most frequently employed models for ultrasound segmentation~\cite{wu2025review}.

\subsection{Brachial Plexus Segmentation: Early Work and Datasets}

Research on deep learning for brachial plexus segmentation gained momentum with the 2016 Kaggle Ultrasound Nerve Segmentation challenge~\cite{kaggle2016uns}, which released 11,143 ultrasound images annotated by clinical experts. Wang et al.~\cite{wang2019resunet} proposed an optimized Residual U-Net (ResU-Net) with residual connections and multi-scale feature preservation, achieving a Dice score of 70.93\% on this dataset. This period established U-Net-based models as strong baselines with Dice scores in the 0.70 range.

\subsection{Architectural Innovations and Multi-Class Extensions}

Subsequent research focused on improving segmentation accuracy through attention mechanisms and contextual modeling. Wu et al.~\cite{wu2021context} proposed a region-aware global context modeling network that captures broader anatomical context around the nerve region, achieving a Dice score of 74.23\% on the Kaggle dataset. The Attention U-Net variant introduced by Oktay et al.~\cite{oktay2018attentionunet} added attention gates that learn to suppress irrelevant regions and highlight salient features, proving effective for small target segmentation in noisy ultrasound images.

A significant advancement was the extension to multi-class segmentation. Ding et al.~\cite{ding2022mallesnet} released the Ultrasound Brachial Plexus Dataset (UBPD), containing 1,055 images with pixel annotations for four classes: nerve, artery, vein, and muscle. They developed MallesNet, which extends Mask R-CNN with spatial local contrast modules and self-attention gates for multi-object segmentation. MallesNet achieved an average Dice of 68.9\% across the four classes, demonstrating the feasibility of simultaneously segmenting multiple anatomical structures. However, the lower per-class accuracy compared to binary segmentation highlighted the increased difficulty of multi-class learning due to class imbalance and inter-class boundary ambiguity.

D{\'\i}az-Vargas et al.~\cite{diazvargas2021cunet} addressed multi-nerve segmentation by developing C-UNet, a conditional U-Net that incorporates nerve-type information via one-hot encoding at the network's deepest layer. This approach enables a single model to segment different peripheral nerves (brachial plexus, sciatic, ulnar, median, femoral) by conditioning on the target structure, achieving approximately 70\% Dice across nerve types. Concurrently, efforts toward real-time deployment led to comparisons with lightweight architectures. LinkNet~\cite{chaurasia2017linknet} was shown to process ultrasound images at 142 frames per second while achieving competitive accuracy (IoU of 66.3\%), suggesting that optimized models could enable real-time clinical guidance~\cite{linknet2022comparison}.

\subsection{Clinical Validation and Cross-Device Generalization}

Recent work has emphasized clinical validation and generalization across imaging devices. Wang et al.~\cite{wang2024bpsegsys} introduced BPSegData, a dataset of 320 brachial plexus trunk images from two different ultrasound machines (185 from one manufacturer, 135 from another). Their BPSegSys system, based on Attention U-Net, achieved a Dice score of 66.87\% and demonstrated that AI performance was comparable to experienced anesthesiologists on the same images. Importantly, this work revealed that cross-device variability contributes significantly to performance degradation, motivating research on domain adaptation and robust training strategies.

Xi et al.~\cite{xi2024cnn} conducted a prospective clinical study in which a U-Net model was trained and tested on 502 interscalene brachial plexus images from 127 patients during actual nerve block procedures. Their model achieved a mean Dice score of 0.748, approaching the threshold they defined for ``good'' segmentation (Dice $> 0.75$). This represents one of the first demonstrations of deep learning segmentation validated in a live clinical context.

More recently, Zhang et al.~\cite{zhang2026labpnet} proposed LA-BPNet, which incorporates anatomical priors through a dual-branch attention mechanism and an Anatomical Information Guidance module that leverages spatial relationships between adjacent nerve roots. By jointly predicting nerve center-points and segmentation masks, LA-BPNet achieved Dice scores exceeding 0.80. A systematic review by Wu et al.~\cite{wu2025review} reported Dice scores ranging from 0.5865 to 0.882 across studies, with U-Net remaining the most frequently employed backbone. The review identified two persistent challenges: scarcity of large, diverse annotated datasets, and poor model generalization across different ultrasound machines or patient populations.

\subsection{Research Gaps}

Despite substantial progress, several gaps remain. First, most studies train and evaluate on data from a single ultrasound device, leaving unclear how to optimally combine heterogeneous data sources. Second, while multi-class segmentation can provide anatomical context, the trade-offs between multi-class and binary supervision for nerve-specific accuracy have not been systematically evaluated. Third, few studies analyze what anatomical factors (such as nerve size) influence segmentation difficulty, which could inform targeted improvements. This study addresses these gaps by examining dataset composition, annotation strategy, and anatomical correlates of segmentation performance under realistic clinical data constraints.

\section{Methodology}
\label{sec:methodology}

\subsection{Dataset}
\label{subsec:dataset}

We utilized a public dataset consisting of ultrasound images of the brachial plexus \cite{ref15}. The data were collected from 101 patients by anesthesiologists using two different ultrasound machines: the SIEMENS ACUSON NX3 Elite and the Philips EPIQ5. Following standard clinical practice, the ultrasound probe was placed at the center of the right side of the neck and slowly moved toward the shoulder. Recordings were captured at a fixed imaging depth of 4~cm with a duration of 8~seconds per patient.

\begin{figure}[ht]
	\centering
	\includegraphics[width=\columnwidth]{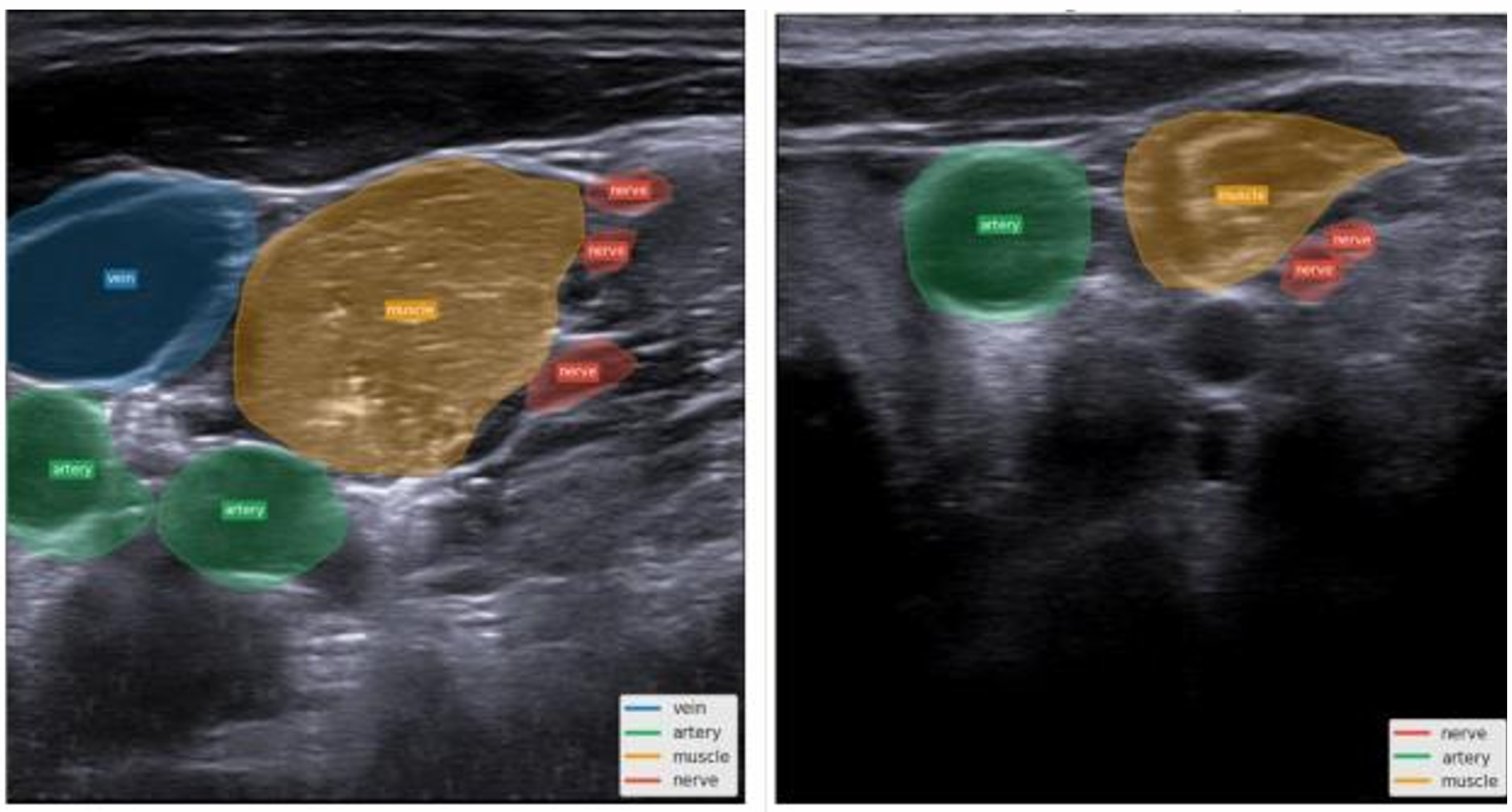}
	\caption{Representative ultrasound images of the brachial plexus with expert annotations for artery, vein, nerve, and muscle.}
	\label{fig:dataset_samples}
\end{figure}

From the 101 recordings, 10--15 frames were randomly selected from each, resulting in a total of 1,052 images. Each image was annotated by an anesthesiologist using LabelMe software, identifying four anatomical structures: artery, vein, nerve, and muscle. The nerve is characterized by an approximate diameter of 3~mm, featuring a bright outer boundary and a darker interior region. Representative images and annotations are shown in \cref{fig:dataset_samples}.

\begin{figure}[ht]
	\centering
	\includegraphics[width=\columnwidth]{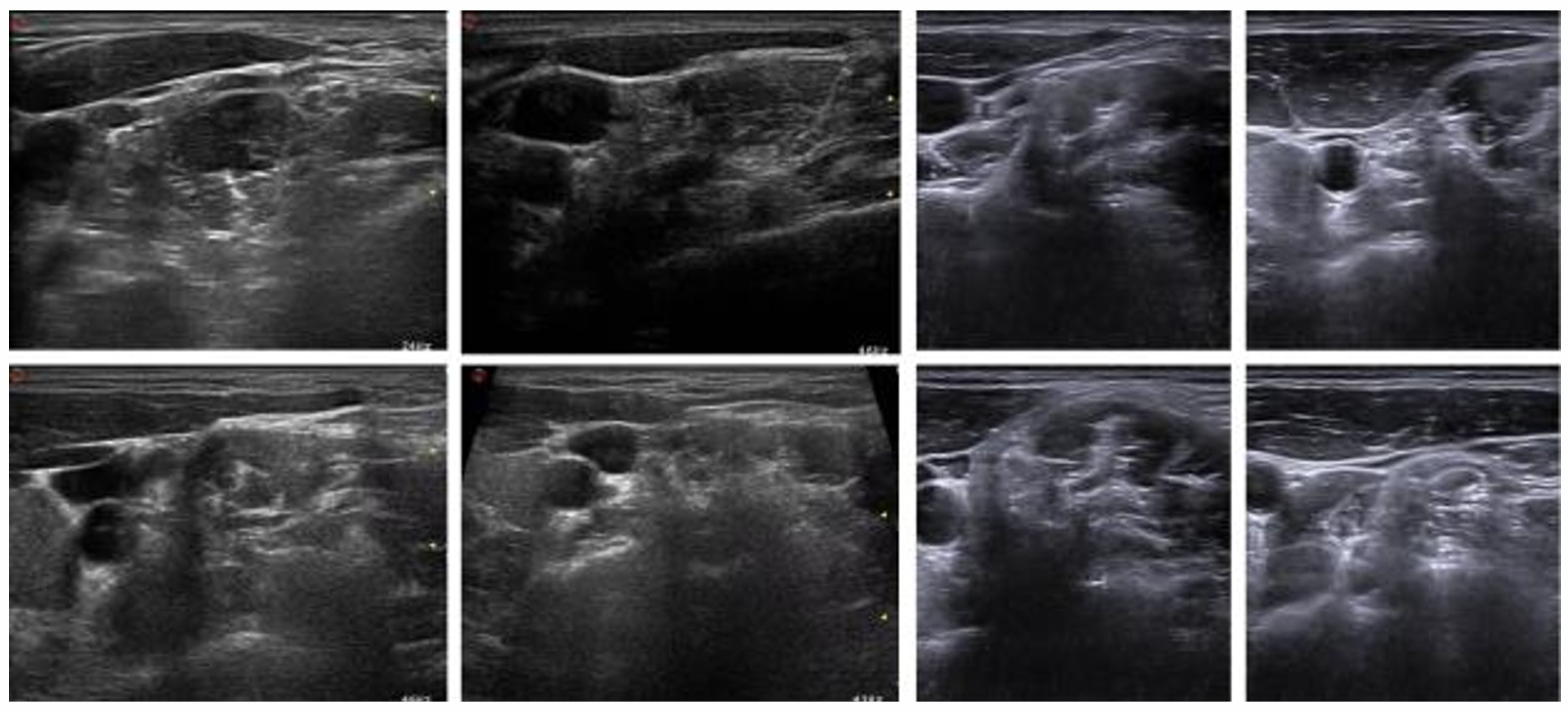}
	\caption{Comparison of ultrasound images from (a) ultrasound-1 (SIEMENS ACUSON NX3 Elite) and (b) ultrasound-2 (Philips EPIQ5).}
	\label{fig:ultrasound_comparison}
\end{figure}

While the original study mentions both ultrasound models, the publicly available dataset contains 955 images and lacks device-specific metadata. To address this, we manually examined the images and classified them into two groups based on visual characteristics: ultrasound-1 (594 images from 52 patients) and ultrasound-2 (361 images from 36 patients). Sample images from both machines are provided in \cref{fig:ultrasound_comparison}.

To characterize the differences between the two machines, we analyzed four image metrics: brightness, contrast, tonal richness, and sharpness. Brightness was measured as the mean pixel value, while contrast was computed as the standard deviation of pixel intensities. Tonal richness was quantified using Shannon entropy, and sharpness was assessed via Laplacian variance. These metrics were calculated for each image as follows:

\begin{equation}
	\text{Brightness} = \frac{1}{N} \sum_{i=1}^{N} I_i
\end{equation}

\begin{equation}
	\text{Contrast} = \sqrt{\frac{1}{N} \sum_{i=1}^{N} (I_i - \mu)^2}
\end{equation}

\begin{equation}
	\text{Sharpness} = \frac{1}{N} \sum_{i=1}^{N} \left| \nabla^2 I_i \right|^2
\end{equation}

\begin{equation}
	\text{Tonal Richness} = -\sum_{k} p_k \log_2 p_k
\end{equation}

where $I_i$ denotes the pixel value, $\mu$ is the mean intensity, $p_k$ is the probability of intensity level $k$, and $\nabla^2$ is the Laplacian operator.

\begin{figure}[ht]
	\centering
	\includegraphics[width=\columnwidth]{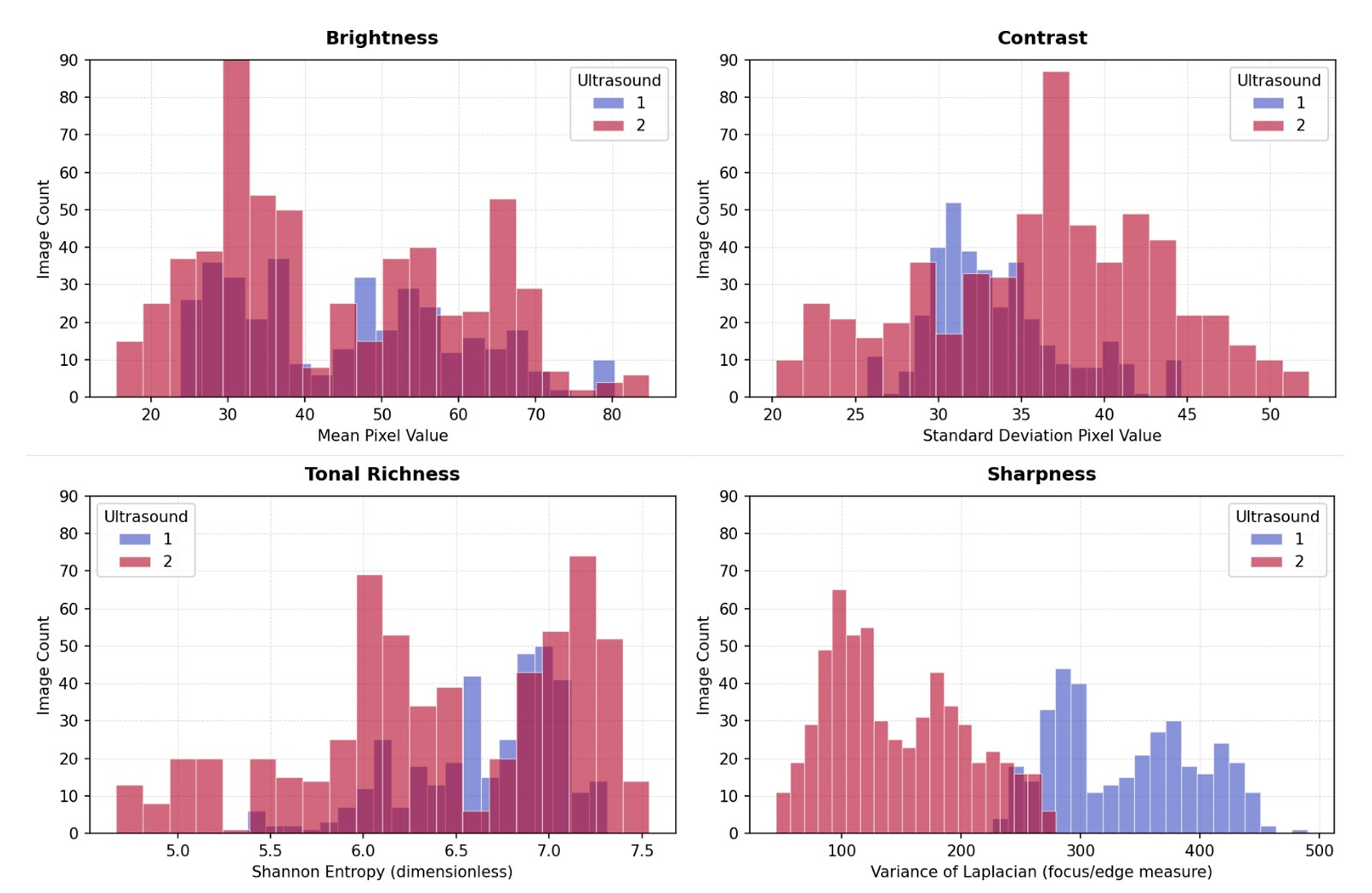}
	\caption{Distribution of image metrics (brightness, contrast, tonal richness, and sharpness) across the two ultrasound machines.}
	\label{fig:metrics_dist}
\end{figure}

The distribution of these metrics is shown in \cref{fig:metrics_dist}.

Brightness, contrast, and tonal richness are relatively consistent across both machines. However, sharpness exhibits a substantial difference: ultrasound-2 shows lower sharpness values compared to the consistently higher variance in ultrasound-1. This indicates that ultrasound-1 captures finer structural details, while ultrasound-2 yields softer boundaries, a factor likely contributing to observed segmentation performance gaps.

\subsection{Data Preprocessing}
\label{subsec:preprocessing}

To prevent data leakage and ensure unbiased evaluation, all datasets were split at the patient level; images from the same patient were never shared across training, validation, or test sets. Input images were then preprocessed to ensure consistency through cropping, resizing, and grayscale conversion.

\paragraph{Cropping}
Cropping was applied to remove irrelevant regions such as text overlays, borders, and patient identifiers. Such metadata can introduce noise and reduce segmentation accuracy, as noted in prior studies \cite{ref9, ref16}. Fixed coordinates were used as all images share a consistent layout.

\paragraph{Resizing}
Following cropping, images were resized by a factor of three to standardize resolution. This ensures uniform input dimensions for the U-Net model, simplifies batch processing, and reduces computational overhead.

\paragraph{Grayscale Conversion}
All images were converted to grayscale. Since ultrasound data are inherently monochromatic, this eliminates redundant color information and reduces input dimensionality, allowing the model to focus on relevant structural patterns.

\subsection{Data Augmentation}
\label{subsec:augmentation}

To increase dataset variability and simulate clinical scenarios, we applied data augmentation techniques including random rotation within $\pm 10^\circ$, translation along the $x$- and $y$-axes up to 12.5\% of image dimensions, and random zooming within a scale range of 0.85--1.25. These augmentations simulate realistic probe motion and imaging depth variation commonly observed during ultrasound acquisition while preserving anatomical integrity.

\subsection{Model Selection}
\label{subsec:model}

\begin{figure}[ht]
	\centering
	\includegraphics[width=\columnwidth]{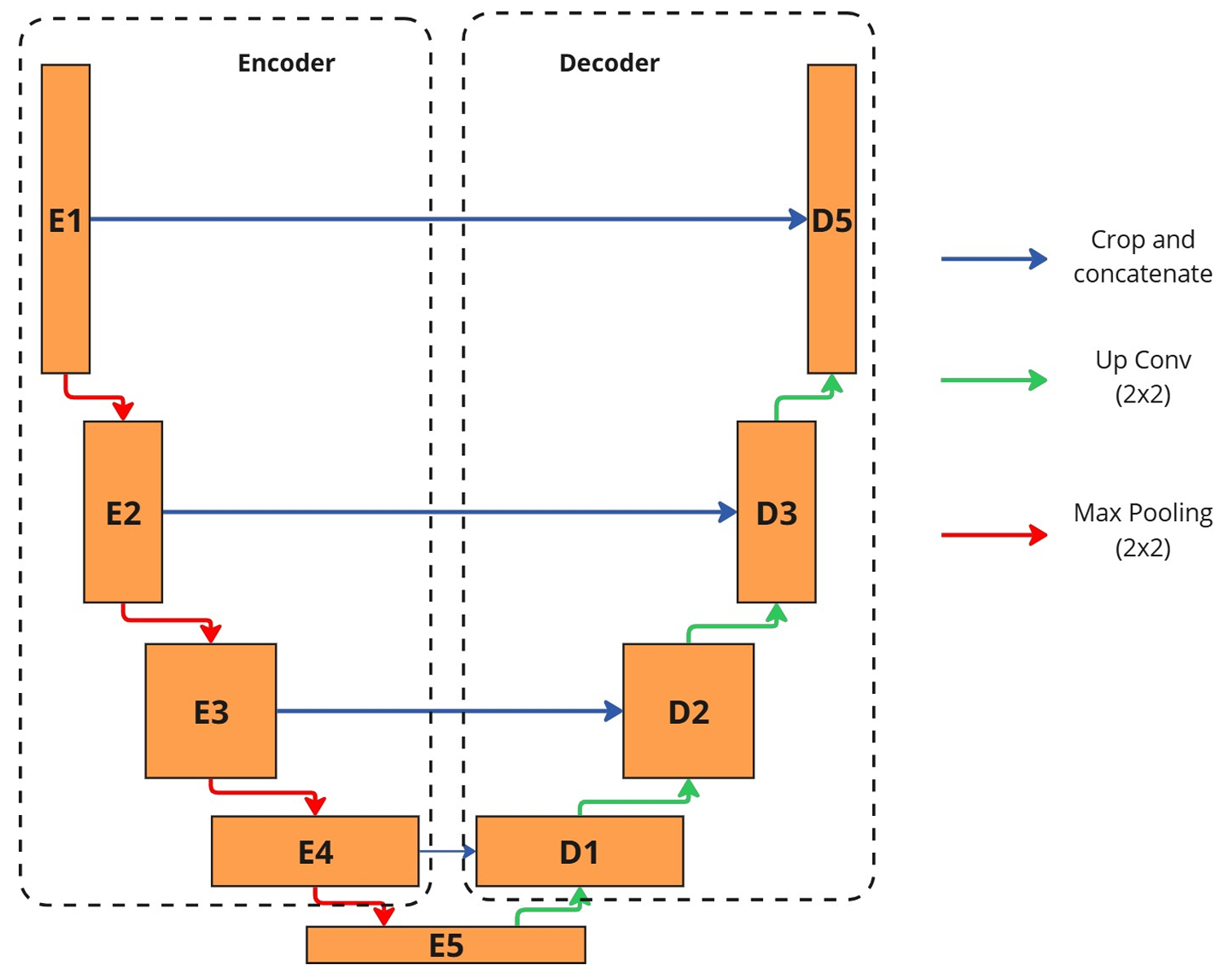}
	\caption{The U-Net architecture used for nerve segmentation, featuring skip connections for boundary preservation.}
	\label{fig:unet_arch}
\end{figure}

We selected the U-Net architecture \cite{ref3} due to its proven effectiveness in medical image segmentation with limited training data. As shown in \cref{fig:unet_arch}, its encoder-decoder structure captures both global context and fine-grained details.

Skip connections help preserve boundary details that are often lost during downsampling, leading to more accurate localization of anatomical landmarks. Rather than optimizing for state-of-the-art performance, U-Net was selected as a stable baseline to ensure that observed performance differences can be attributed to data and training strategy rather than architecture.

\subsection{Training Configuration}
\label{subsec:training}

All models were trained using a supervised learning framework in PyTorch. The loss function was selected based on annotation strategy: Binary Cross-Entropy (BCE) loss for binary segmentation involving a single anatomical landmark, and Categorical Cross-Entropy (CCE) loss for multi-class segmentation involving multiple landmarks.

Model optimization was performed using the Adam optimizer with an initial learning rate of $1 \times 10^{-3}$. Performance was evaluated using five-fold cross-validation with GroupKFold to ensure patient-level splitting. To mitigate overfitting, early stopping was employed with a patience of 10 epochs based on validation performance. A ReduceLROnPlateau scheduler adjusted the learning rate when validation performance stagnated. A fixed batch size of 32 was used across all experiments. All models were trained using half-precision (FP16) to improve efficiency while maintaining numerical stability. Training was performed on a single NVIDIA Tesla V100 GPU with 16~GB memory.

\subsection{Experiment Design}
\label{subsec:experiments}

Two experiments were conducted using the brachial plexus dataset to investigate the influence of dataset composition and annotation strategy.

\begin{figure}[ht]
	\centering
	\includegraphics[width=\columnwidth]{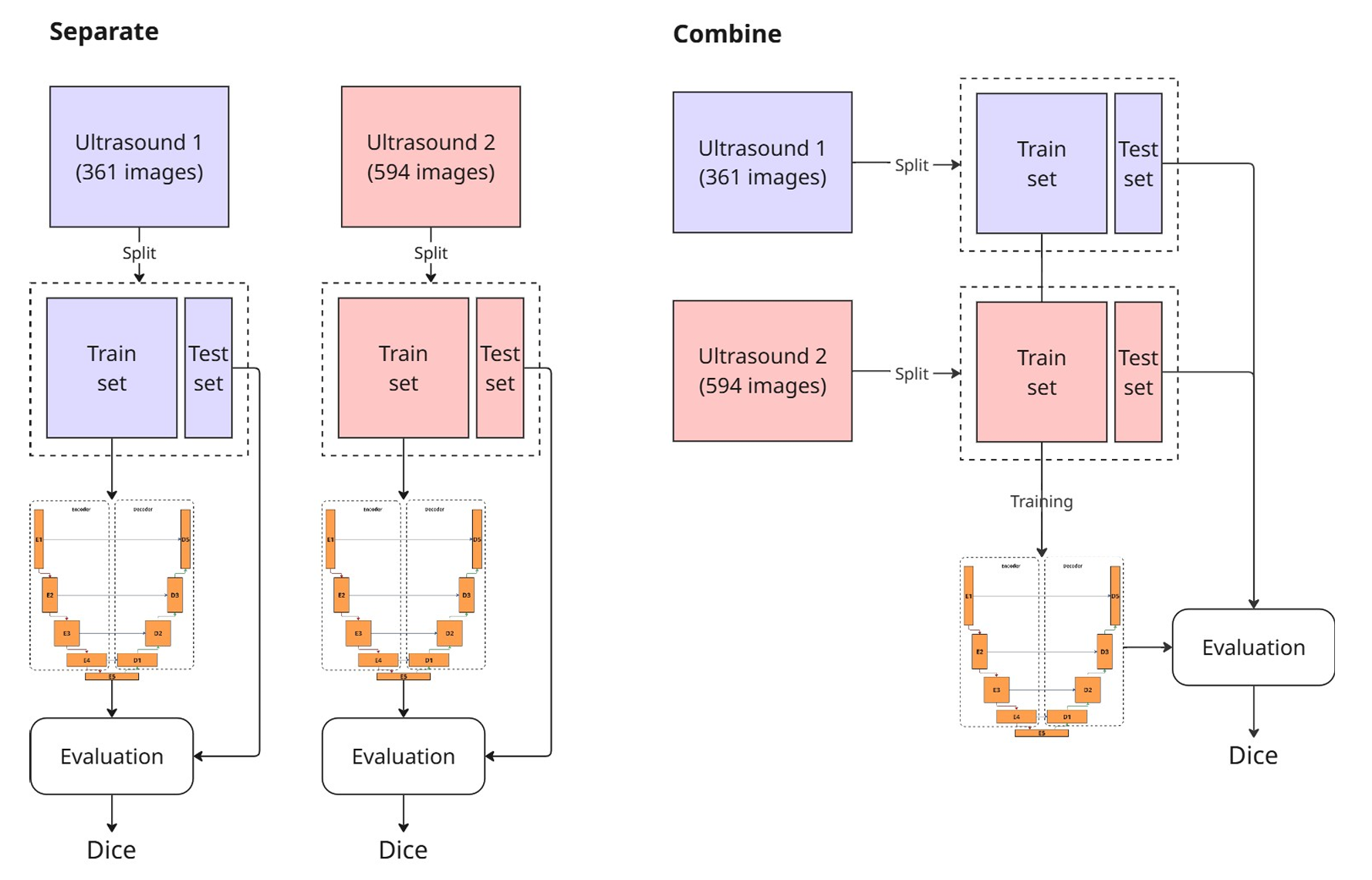}
	\caption{Experimental setup for combined dataset training and cross-machine evaluation.}
	\label{fig:exp_setup}
\end{figure}

\paragraph{Experiment 1: Dataset Composition}
This experiment examined whether combining data from different ultrasound machines improves segmentation performance. Models were trained separately on ultrasound-1 (594 images), ultrasound-2 (361 images), and a combined dataset (955 images). Performance was cross-evaluated as illustrated in \cref{fig:exp_setup}.

\paragraph{Experiment 2: Multi-class Annotation}
We investigated whether including additional anatomical landmarks (artery, vein, and muscle) improves nerve segmentation. Multi-class models were compared against binary models (nerve only) to evaluate whether structural context assists in identifying ambiguous nerve regions.

\subsection{Evaluation Metric}
\label{subsec:evaluation}

The system was evaluated using the Dice score, a standard metric for medical image segmentation that measures spatial overlap between the predicted mask $P$ and ground-truth $G$ \cite{ref10}:
\begin{equation}
	\text{Dice}(P, G) = \frac{2|P \cap G|}{|P| + |G|}
\end{equation}

\section{Results and Discussion}
\label{sec:results}

We evaluated the performance of our segmentation model across several experimental configurations to understand the impact of dataset diversity and supervision strategy. For each configuration, we performed train-test splits using three distinct random seeds (33, 42, and 94) to ensure statistical robustness and conducted cross-validation to obtain reliable performance estimates.

\subsection{Dataset Composition Analysis}
\label{subsec:dataset_composition}

Our first experiment examined whether combining data from multiple ultrasound machines improves generalizability. We compared performance when training on isolated datasets (Ultrasound-1 and Ultrasound-2) versus a combined dataset.

As shown in \cref{tab:table_nerve_only}, models trained on U2 alone achieved the highest mean Dice score (0.407), followed by the combined dataset (0.388), while U1-only training performed worst (0.274). Notably, U2 images exhibit lower sharpness than U1 (\cref{fig:metrics_dist}), suggesting that the higher U2 performance may reflect greater consistency within that dataset rather than inherent image quality advantages.

The combined dataset did not surpass U2 in absolute performance, but it substantially improved over U1-only training (+0.114 Dice). This indicates that incorporating diverse acquisition sources helps models trained on lower-performing data benefit from cross-machine regularization. The combined model thus offers a practical trade-off: slightly lower peak performance than the best single-source model, but more robust generalization when deployment conditions are uncertain.

\begin{table}[t]
	\caption{Mean Dice scores for binary nerve segmentation across dataset configurations. Results are averaged over three random seeds (33, 42, 94) with standard deviation shown.}
	\label{tab:table_nerve_only}
	\vskip 0.15in
	\begin{center}
		\begin{small}
			\begin{sc}
				\begin{tabular}{lc}
					\toprule
					Training Data      & Mean Dice         \\
					\midrule
					U1 only            & $0.274 \pm 0.087$ \\
					U2 only            & $0.407 \pm 0.124$ \\
					Combined (U1 + U2) & $0.388 \pm 0.094$ \\
					\bottomrule
				\end{tabular}
			\end{sc}
		\end{small}
	\end{center}
	\vskip -0.1in
\end{table}

\subsection{Multi-landmark Segmentation Challenges}
\label{subsec:multi_class}

We further extended the segmentation task to include multiple anatomical landmarks: artery, vein, nerve, and muscle. Surprisingly, the addition of these classes led to a substantial drop in the Dice score for the nerve class across all configurations, as shown in \cref{tab:table_multi_class}.

\begin{figure}[ht]
	\centering
	\includegraphics[width=\columnwidth]{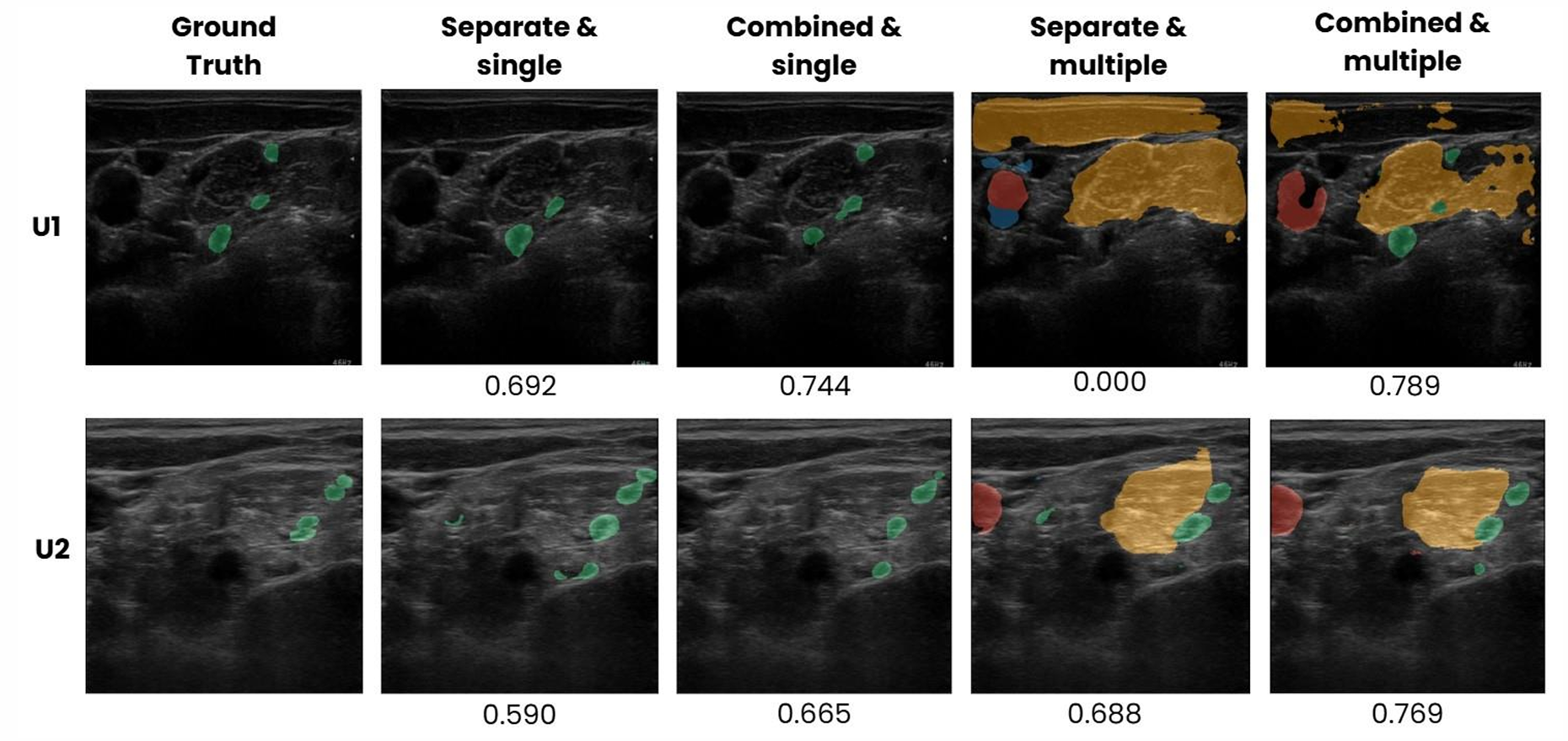}
	\caption{Representative segmentation results showing input ultrasound images, ground-truth annotations, and model predictions.}
	\label{fig:segmentation_results}
\end{figure}

Although additional anatomical supervision often aids in contextual understanding, the observed decline suggests that multi-class learning introduces significant complexities for this specific task. The most severe degradation occurred with U1 training, where nerve Dice dropped from 0.274 to 0.107, a 61\% relative decrease. U2 and combined datasets showed more modest declines of 11\% and 9\%, respectively. These challenges may stem from severe class imbalance between the nerve and surrounding muscle tissue, competition for shared representational capacity, and visual similarity between soft tissue structures that introduces ambiguity at nerve boundaries. Qualitative segmentation results are provided in \cref{fig:segmentation_results}, showcasing representative model predictions.

\begin{table}[t]
	\caption{Comparison of nerve Dice scores under binary versus multi-class supervision. Multi-class training degrades nerve segmentation across all configurations, with U1 showing the largest drop.}
	\label{tab:table_multi_class}
	\vskip 0.15in
	\begin{center}
		\begin{small}
			\begin{sc}
				\begin{tabular}{lccc}
					\toprule
					Training Data & Binary & Multi-class & Change  \\
					\midrule
					U1 only       & 0.274  & 0.107       & $-$61\% \\
					U2 only       & 0.407  & 0.363       & $-$11\% \\
					Combined      & 0.388  & 0.355       & $-$9\%  \\
					\bottomrule
				\end{tabular}
			\end{sc}
		\end{small}
	\end{center}
	\vskip -0.1in
\end{table}

\subsection{Nerve Size and Segmentation Accuracy}
\label{subsec:nerve_size}

To examine the relationship between nerve size and segmentation performance, nerve size was quantified from ground-truth binary masks as the total number of nerve pixels:
\begin{equation}
	\text{Nerve Size} = \sum_{i=1}^{N} M_i
\end{equation}
where $M_i \in \{0,1\}$ denotes the mask value at pixel $i$, and $N$ is the total pixel count.

Using Dice scores aggregated across all experiments, a moderate positive correlation was observed between nerve size and segmentation accuracy (Pearson $r=0.587$, $p=9.0 \times 10^{-6}$). This indicates that larger nerves tend to be segmented more reliably, while smaller nerves remain more challenging for the model.

This trend can be attributed to the limited pixel representation of small nerves, which makes boundary delineation more sensitive to prediction errors. For small anatomical structures, even minor spatial deviations lead to substantial Dice score reduction. In contrast, larger nerves provide stronger structural cues and greater spatial redundancy, allowing models to achieve higher overlap with ground truth. These findings highlight nerve size as an important factor influencing segmentation performance and help explain variability observed across samples.

\section{Conclusion}
\label{sec:conclusion}

This study evaluated deep learning-based nerve segmentation in ultrasound images of the brachial plexus, examining how dataset composition and annotation strategy influence model performance. Our results show that while U2-only training achieved the highest absolute Dice score (0.407), training on combined data from multiple ultrasound machines substantially improved performance over U1-only training (0.388 vs.\ 0.274). This suggests that dataset diversity provides regularization benefits, particularly for lower-performing acquisition sources, even when it does not surpass the best single-source performance.

Extending from binary nerve segmentation to multi-land\-mark supervision resulted in decreased nerve Dice scores across all configurations, with U1 experiencing a 61\% relative drop. This suggests that without careful class balancing or weighting, additional anatomical supervision may introduce competition between classes and degrade performance on clinically critical structures. Our correlation analysis confirmed that larger nerves are segmented more reliably (Pearson $r=0.587$, $p < 0.001$), while smaller nerves remain challenging due to limited pixel representation and sensitivity to prediction errors.

Overall, these results demonstrate that effective ultrasound nerve segmentation depends critically on training data composition and learning strategy. When deployment conditions are uncertain, combined training offers a robust compromise; when target acquisition characteristics are known, single-source training on matched data may be preferable.

\section{Future Work}
\label{sec:future}

The observed performance degradation under multi-land\-mark supervision motivates detailed ablation studies of anatomical context. Rather than jointly training on all landmarks, future experiments could evaluate selective combinations (nerve with artery, nerve with vein, or nerve with muscle) to identify which structures provide beneficial contextual cues and which introduce detrimental class competition.

Although GPU-based inference achieved acceptable performance, CPU-only inference remained insufficient for practical deployment. Future work should focus on optimizing inference speed through model compression, architectural simplification, or quantization, which are critical for enabling deployment in point-of-care or resource-constrained clinical environments where dedicated GPU hardware may not be available.

Beyond the scope of this study, future research could explore temporal modeling across video frames, uncertainty-aware pseudo-label selection, and prospective clinical validation. Incorporating temporal consistency or human-in-the-loop feedback may further improve robustness and facilitate integration into clinical ultrasound workflows.

% Acknowledgements should only appear in the accepted version.
% \section*{Acknowledgements}

% \textbf{Do not} include acknowledgements in the initial version of the paper
% submitted for blind review.

% If a paper is accepted, the final camera-ready version can (and usually should)
% include acknowledgements.  Such acknowledgements should be placed at the end of
% the section, in an unnumbered section that does not count towards the paper
% page limit. Typically, this will include thanks to reviewers who gave useful
% comments, to colleagues who contributed to the ideas, and to funding agencies
% and corporate sponsors that provided financial support.

\section*{Acknowledgements}

We extend our gratitude to the Innovative Cognitive Computing Research Lab for providing workspace support.

\section*{Impact Statement}

This paper presents work whose goal is to advance the field of Machine
Learning. There are many potential societal consequences of our work, none
which we feel must be specifically highlighted here.

% In the unusual situation where you want a paper to appear in the
% references without citing it in the main text, use \nocite
% \nocite{langley00}

\bibliography{main}
\bibliographystyle{icml2026}

%%%%%%%%%%%%%%%%%%%%%%%%%%%%%%%%%%%%%%%%%%%%%%%%%%%%%%%%%%%%%%%%%%%%%%%%%%%%%%%
%%%%%%%%%%%%%%%%%%%%%%%%%%%%%%%%%%%%%%%%%%%%%%%%%%%%%%%%%%%%%%%%%%%%%%%%%%%%%%%
% APPENDIX
%%%%%%%%%%%%%%%%%%%%%%%%%%%%%%%%%%%%%%%%%%%%%%%%%%%%%%%%%%%%%%%%%%%%%%%%%%%%%%%
%%%%%%%%%%%%%%%%%%%%%%%%%%%%%%%%%%%%%%%%%%%%%%%%%%%%%%%%%%%%%%%%%%%%%%%%%%%%%%%
% \newpage
% \appendix
% \onecolumn
% \section{You \emph{can} have an appendix here.}

% You can have as much text here as you want. The main body must be at most $8$
% pages long. For the final version, one more page can be added. If you want, you
% can use an appendix like this one.

% The $\mathtt{\backslash onecolumn}$ command above can be kept in place if you
% prefer a one-column appendix, or can be removed if you prefer a two-column
% appendix.  Apart from this possible change, the style (font size, spacing,
% margins, page numbering, etc.) should be kept the same as the main body.
% %%%%%%%%%%%%%%%%%%%%%%%%%%%%%%%%%%%%%%%%%%%%%%%%%%%%%%%%%%%%%%%%%%%%%%%%%%%%%%%
% %%%%%%%%%%%%%%%%%%%%%%%%%%%%%%%%%%%%%%%%%%%%%%%%%%%%%%%%%%%%%%%%%%%%%%%%%%%%%%%

\end{document}